\documentclass[journal]{IEEEtran}
\usepackage{times}

\usepackage[numbers]{natbib}
\usepackage{multicol}
\usepackage[bookmarks=true]{hyperref}
\usepackage{subcaption}
\usepackage{graphicx} 
\usepackage{epsfig} 
\usepackage{mathptmx} 
\usepackage{amsmath} 
\usepackage{amssymb}  
\usepackage{bm}
\usepackage{array}
\usepackage{balance}
\newcolumntype{P}[1]{>{\centering\arraybackslash}p{#1}}
\usepackage{xcolor}
\usepackage{hyperref}
\usepackage{footnote}
\usepackage{bigints}
\usepackage{url}
\hypersetup{
    colorlinks,
    citecolor=blue,
    filecolor=blue,
    linkcolor=blue,
    urlcolor=blue
}

\begin{document}

\title{\emph{ROSE}: Rotation-based Squeezing Robotic Gripper toward Universal Handling of Objects}

\author{Son Tien Bui$^{1}$,
        Shinya Kawano$^{2}$,
        and Van Anh Ho$^{3}$ \\
        \emph{Soft Haptics Lab, Japan Advanced Institute of Science and Technology (JAIST)}\\
        
        Email: {\tt\small  van-ho@jaist.ac.jp}
\thanks{$^{1}$ Bui is with Hanoi University of Industry from April $1^{st}$, 2023, Email: sonbt@haui.edu.vn}
\thanks{$^{2}$ Kawano is with Mitsubishi Electrics Corp. from April $1^{st}$, 2023, Email: kawano.shinya.serious@gmail.com}
\thanks{$^{3}$ Ho is also with Japan Science and Technology Agency, PRESTO, Kawaguchi Saitama 332-0012 Japan.}}


\maketitle

\begin{abstract}
Robotics hand/grippers nowadays are not limited to manufacturing lines; instead, they are widely utilized in cluttered environments, such as restaurants, farms, and warehouses. In such scenarios, they need to deal with high uncertainty of the grasped objects' shapes, postures, surfaces, and material properties, which requires complex integration of sensing and decision-making process. On the other hand, integrating soft materials into the gripper's design may tolerate the above uncertainties and reduce complexity in control. In this paper, we introduce \emph{ROSE}, a novel soft gripper that can embrace the object and squeeze it by buckling a funnel-liked thin-walled soft membrane around the object by simple rotation of the base. Thanks to this design, ROSE hand can adapt to a wide range of objects that can {fit in the} funnel and handle with {gentle} gripping force. Regardless of this, ROSE can generate a high lift force (up to 33\,kgf) while significantly reducing the normal pressure on the gripped objects. In our experiment, a 198\,g ROSE can be integrated into a robot arm with a single actuation and successfully lift various types of objects, even after 400,000 trials. The embracing mechanism helps reduce the dependence of friction between the object and the membrane, as ROSE could pick up a chicken egg submerged inside an olive oil tank. We also report a feasible design for equipping the ROSE hand with tactile sensing while appealing to the scalability of the design to fit a wide range of objects.\\
Video: \url{https://youtu.be/E1wAI09LaoY}
\end{abstract}

\IEEEpeerreviewmaketitle

\section{Introduction}
Robots have contributed to increasing manufacturing productivity and reducing risks for humans in operation in harsh environments. One of the most popular tasks is grasping/manipulation, which results in a great deal of research related to the development of robotic hand/grippers with various mechanisms. Since human hands and other similar organs of nature are dexterous in manipulation while adaptive enough to a wide range of grasped objects; robotics research tends to mimic both structural and functional, even sensing/perception, to design of robotic hands/grippers.

\par Owing to the integration of soft materials, soft grippers increasingly resemble nature's structures. They enable gentle touch, high customizability, and dynamic adaptation to various objects. Inspired by nature, soft grippers often mimic gripping biological mechanisms such as the human hand (fingers and palm \cite{finger_1}-\cite{octopus_suction_1}), locking forms of plans (Venus flytrap \cite{flytrap}) or animals (octopus tentacles \cite{octopus_suction_1, octopus_suction_2}, elephant trunks \cite{elephenttrunk1, elephenttrunk2}, bird perching \cite{perching}). In addition, the opening-closing mechanisms of {flower petals also inspired the design of soft grippers} \cite{flowers1, flowers2}. Most soft robotics hand attempt to squeeze the object to create a large-area contact to increase the stability of the grasping action. Compared to robotic caging \cite{CAGE} using rigid fingered hands that may pose complexity in maintaining contacts, soft robotic hands could embrace objects without implementing complicated control algorithms \cite{WRAP}, thanks to their flexibility and deformability. On top of that, the durability of a soft robotic is of great concern which may prevent a wide usage in actual scenarios \cite{roadmap}.

\begin{figure}
    \centering
    \includegraphics[width =\linewidth]{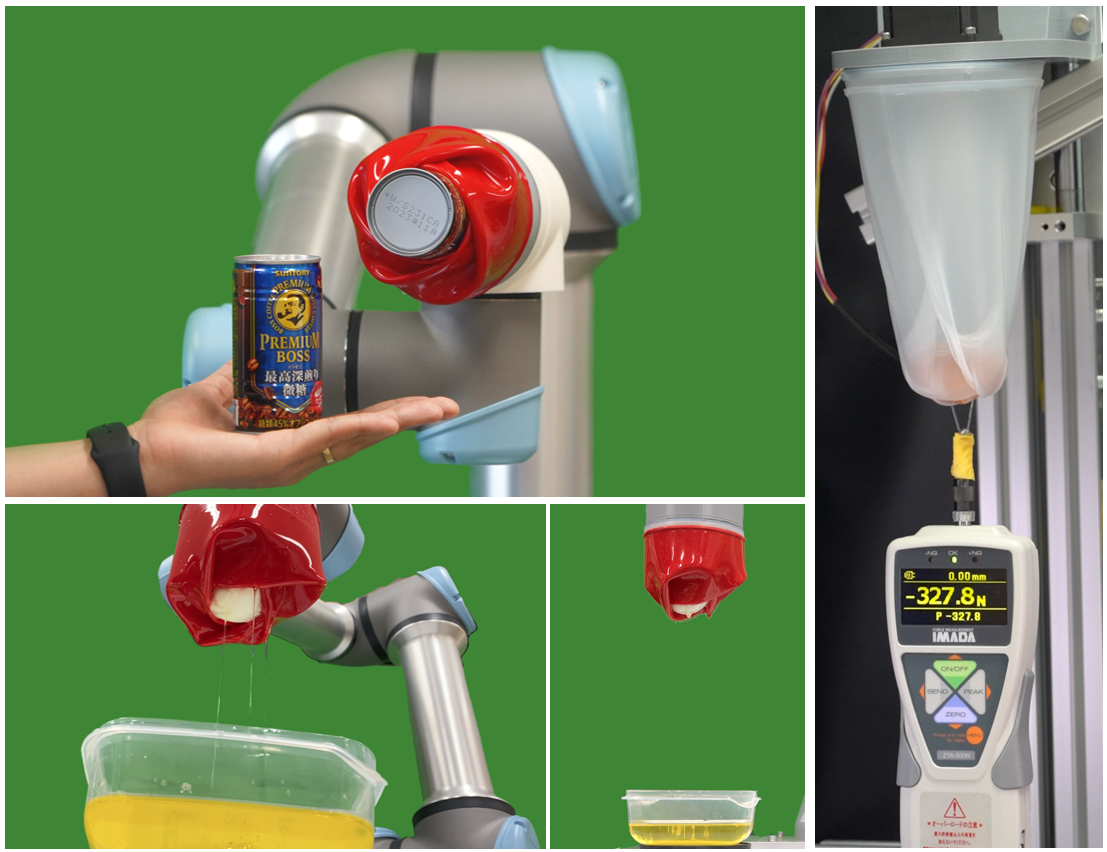}
    \caption{From right to left, top to bottom: \emph{ROSE} gripper on a robot arm picking a coffee can, and a peeled chicken egg submerged inside an olive oil container. The maximum lifting force that ROSE can endure is 328.7\,N.}
    \label{fig: rose_introduction}
\end{figure}
\par In this paper, we present \textbf{ROSE} (\textbf{RO}tation-based \textbf{S}queezing gripp\textbf{E}r). The design took the hint from the blooming states of a rose, promising gentle gripping ability while maintaining the embracement, toward universal handling of objects (see Figure \ref{fig: rose_introduction}). The design of ROSE is simple, whose initial shape resembles a ferrule sleeve, yet yielding high gripping force (up to 33\,kgf), as well as a payload-to-weight ratio (about 6812$\%$ for the gripper alone). With a single rotation, ROSE can squeeze most object that falls inside the region bounded by the ferrule sleeve. The simplicity of mechanical design allows high scalability, ease of fabrication, and high durability. The grippers could endure up to 400,000 open-close trials, which promises {productivity} usage in actual cases. 

The paper includes section \ref{design_fabrication} for explaining the design idea and fabrication process of the novel soft robotic gripper, followed by a modeling approach for estimation of applied pressure on the gripped objects in section \ref{modelling}. In section \ref{experiments}, we report the payload experiment, as well as the durability test, with the preliminary elaboration of the sensing ability, follows by the results in section \ref{results}. Last but not least, section \ref{discussion_and_conlcusion} states discussions on the limitation of the proposed hand, the model, and tactile sensing, followed by future work.

\section{Related Works}
\label{related_works}
Since soft robot hands/grippers are inherently different from rigid ones, in this investigation, we focus on the elaboration of the design and function of two main groups as follows:
\subsection{Fingered Soft Grippers}
Similar to rigid-fingered robotic hands, designs of soft robotic hands also attempt to mimic the structure and function of human hands. Several designs were commercialized\footnote{\url{https://www.softroboticsinc.com/}}, while some challenges in dexterity and durability remained. Liu \emph{et al.} \cite{2fingers} introduces a two-fingered gripper that could grasp objects of different shapes and sizes, from small to fragile ones.  A three-fingered gripper with an active palm, proposed by Pagoli \emph{et al.} \cite{finger_palm}, could conduct in-hand manipulation such as rotating a Rubik's cube. Recently, a four-fingered gripper for caging and lifting various objects was introduced in \cite{4fingers}. Other approaches focused on increasing the gripping function of the fingered hand in specific scenarios. For example, micropatterned pads were deposited to the fingertip's surface of a two-finger gripper developed in \cite{tofugripper} could enhance wet adhesion in gripping wet and deformable objects such as tofu. Despite the difference in structure, these grippers required pneumatic systems with extra valves and sensors, which are not always convenient to set up. In addition, such a fingered hand needs to adjust its posture w.r.t. location and orientation of the objects precisely, resulting in complications in perception and control systems. Last but not least, the normal force applied to the object's surface of the soft finger is inherently low, {which may fall outside the friction cone}, resulting in instability in grasping and manipulation.
\subsection{Large Contact-based Soft Grippers}
Thanks to their deformability, soft structures tend to make large-area contact with the object, thus increasing the stability of grasping. A typical design is a tube-shaped gripper with a negative air support system that generates centripetal force to grip an object \cite{magicball, Haili2023, Li2021b, Pedro2018, 9200339}. As a result of the symmetrical design, these grippers could grasp stably various objects while offering a high payload-to-weigh ratio of up to 120 \cite{magicball}. However, these grippers fail to grasp unsuitable objects such as lying bars or oversize spherical where they cannot be "trapped" inside. Likewise, the unstable shrinkage of the origami structure leads to the change of grasped objects' posture, which frustrates robotic manipulation like the assembly task \cite{magicball}.
\par One method for increasing the contact area is the jamming structure \cite{jamminggripper1, jamming2, seasampling}, which is only applicable to the soft gripper. Here, {the negative pneumatic helps the particles interlock thus increasing its stiffness and the gripping force to the object’s surface}, which significantly reduces the complexity of the gripper's structure and related control regime. Nevertheless, if the soft membranes get torn, the gripper becomes unfunctional. Additionally, this jamming principle requires these grippers to press against the object, which might damage fragile ones (such as fruit or food products).
\par Other grippers combine the ideas of large contact-area to finger designs \cite{softbuble, scooplinggripper} or using special principles such as wrapping \cite{WRAP} or rolling \cite{elephenttrunk1, elephenttrunk2}. Despite the grasping successes, the durability, the in-hand manipulation, and the working ability in cluttered environments of these novel designs remain challenging. 


\subsection{Expected Contributions}
Taking into account the above limitations of current development, we {propose} a novel design of a soft robotic hand that can embrace and grip firmly any object that falls into the gripping region with one rotation. The gripper can accommodate uncertainties of the object's location and orientation while providing gentle touch with high friction to the object. The developed hand was tested with 400,000 trials without failing the gripping function. The main contributions of this paper are as follows:
\begin{enumerate}
    \item Proposal of a design and a feasible fabrication method of ROSE, a soft gripper with rotation-based buckling structure.
    \item Proposal of simplified models for investigating applied pressure on the object upon squeezing by the ROSE gripper.
    \item Proposal of contact perception of ROSE gripper using vision-based tactile sensing technique.
\end{enumerate}
\section{Design and Fabrication}
\label{design_fabrication}
\subsection{Idea}
\begin{figure}[t]
     \centering
     \begin{subfigure}[t]{0.30\linewidth}
        \centering
        \includegraphics[width = \linewidth]{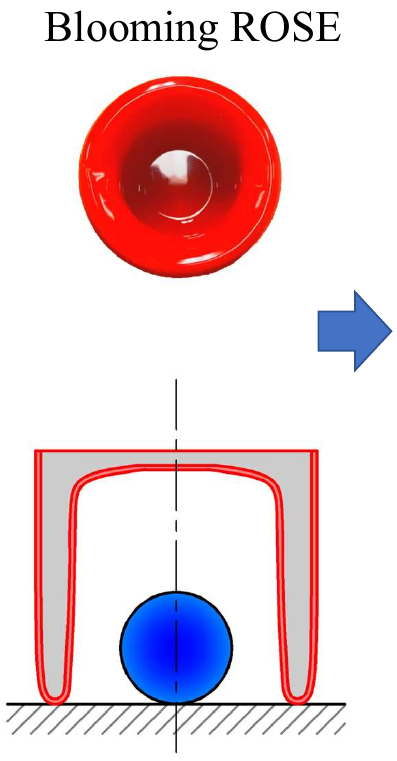}
        \caption{the Approaching}
        \label{fig: rose_approaching}
     \end{subfigure}
     \hfill
     \begin{subfigure}[t]{0.305\linewidth}
         \centering
         \includegraphics[width = \linewidth]{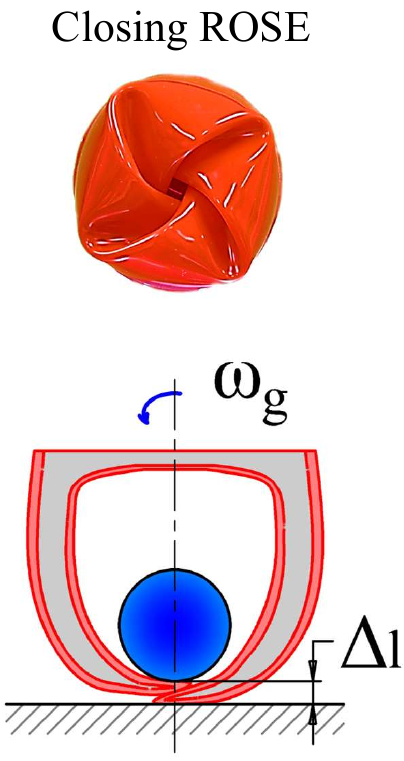}
        \caption{the Lifting}
        \label{fig: rose_lifting}
     \end{subfigure}
     \hfill
     \begin{subfigure}[t]{0.338\linewidth}
         \centering
        \includegraphics[width = \linewidth]{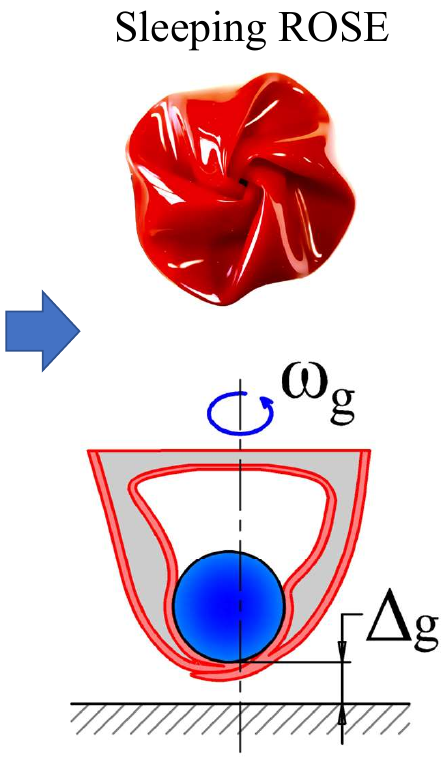}
        \caption{the Holding}
        \label{fig: rose_gripping}
     \end{subfigure}
     \hfill
        \caption{ROSE's gripping process through three main steps: approaching, lifting, and holding an object at the heights of 0, $\Delta_l$, and $\Delta_g$, respectively.}
        \label{fig: rose_gripping_process}
\end{figure}
\begin{figure*}[ht]
     \centering
     \begin{subfigure}[t]{0.38\linewidth}
        \centering
        \includegraphics[width = \linewidth]{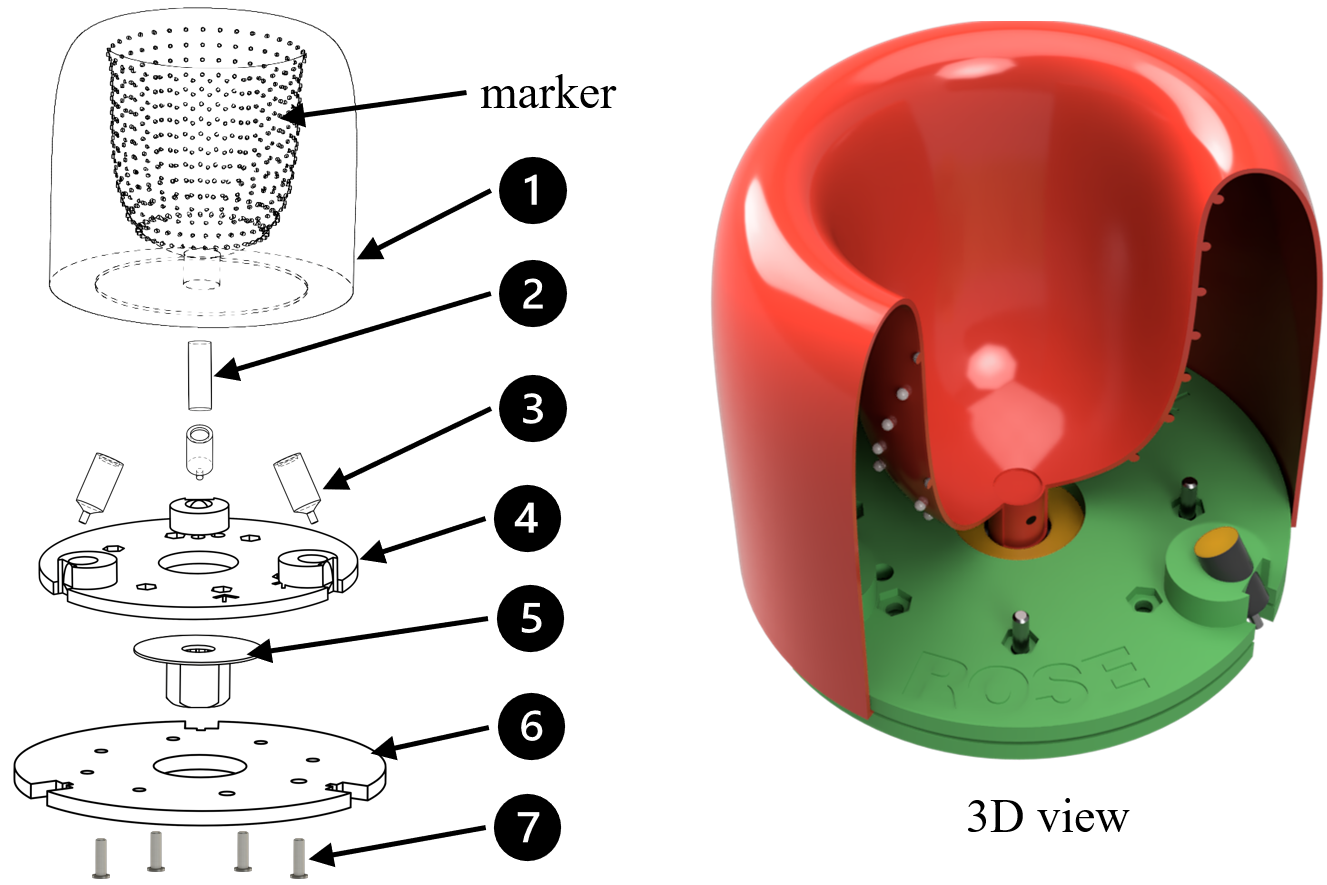}
    \caption{The design of ROSE gripper, from left to right: explore view and 3D view; 1 - skin with markers {(petals)}, 2 - air tube, 3 - cameras, 4 - lower base, 5 - cylinder, 6 - lower base, 7 - bolts}
    \label{rose_design}
     \end{subfigure}
     \hfill
     \begin{subfigure}[t]{0.60\linewidth}
        \centering
        \includegraphics[width = \linewidth]{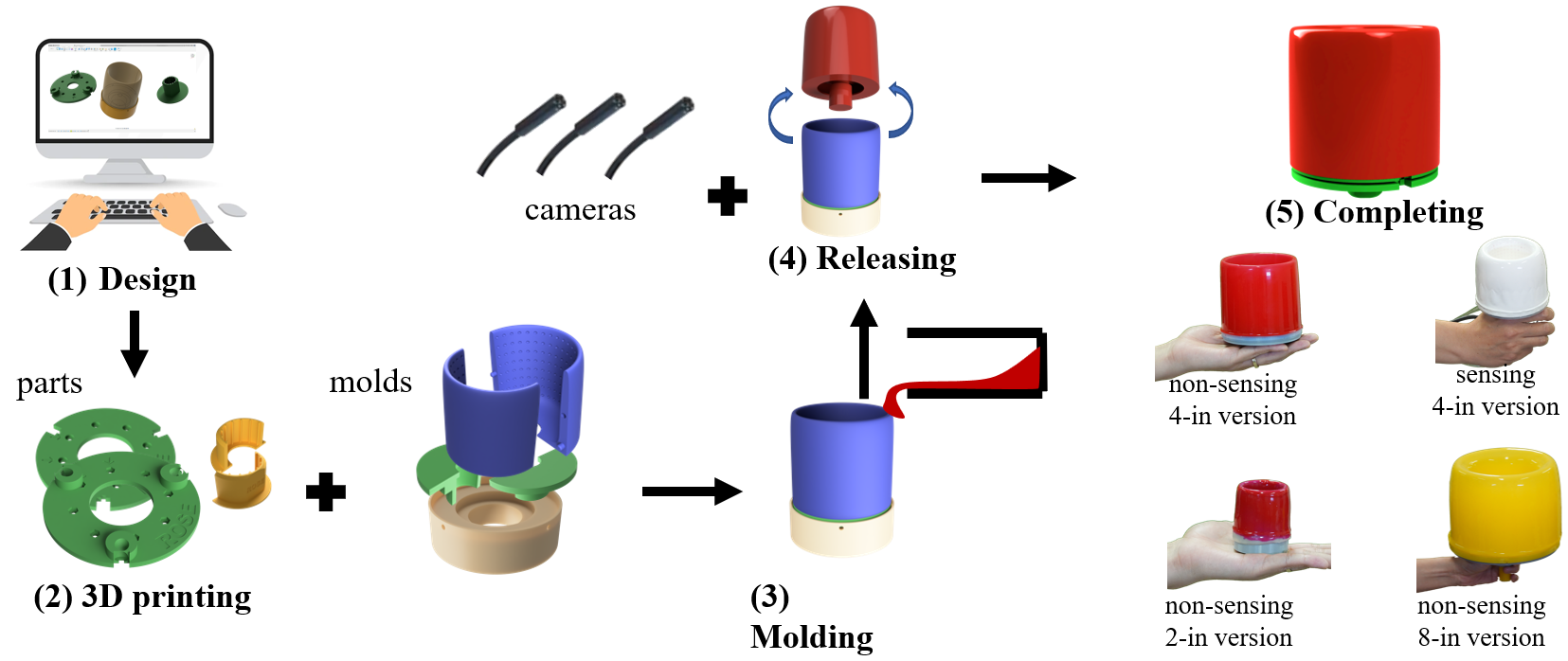}
    \caption{Process for fabrication of the ROSE gripper. Note that the gripper can be scalable using this process.}
    \label{fabrication}
     \end{subfigure}
     \hfill
        \caption{ROSE gripper's design and the skin's fabrication process.}
        \label{fig: design_and_fabrication}
\end{figure*}
ROSE gripper's outer shape during operation is resemble to a fully bloomed rose, while the closing state has a shape similar to a sleeping rose (in dormancy), even though there is no explicit mimicking here. If any object falls in the region bounded by the petal, under a twist/rotation-action, ROSE hand can embrace and grip it safely (see illustration in Fig. \ref{fig: rose_approaching}). In such a scenario, ROSE shares a close working principle with traditional granular jamming gripper since the object is covered by a membrane (soft skin) and held by normal pressure. With this idea, the close/open operation of the hand is implemented by a chain of complex buckling deformation of a membrane under a twist action. They can be divided into three steps: Approaching, Lifting, and Holding. First, ROSE approaches the object (see Figure \ref{fig: rose_approaching}) and makes sure the object is within the petal. Next, the central cylindrical base rotates at a rotational speed $\omega_g$ leading to the buckling of the skin, gradually embracing the object (see Figure \ref{fig: rose_lifting}). When the base keeps rotating, it increases the contact area and the applied pressure on the object's surface. Therefore, the generated gripping force on the object's surface would grow, which in turn helps ROSE squeeze the object firmly (see Figure \ref{fig: rose_gripping}). Note that the rotational angles of steps 2 and 3 could be adjusted to adapt to the gripping objects' size, shape, and weight.

\subsection{Design}


Instead of proposing a multi-layered structure consisting of separated petals stitched together \cite{flowers1, flowers2}, we simplified the idea with a symmetrical design and aimed to bring ROSE gripper with vision-based contact sensing in a complete structure. Details of the ROSE's design can be seen in Figure \ref{rose_design}, in which the skin(1) continuously connects the outer ring of the lower base (6) and the center of the upper base (4), designed to avoid concentrated tension at the rotational axis by a transitioning curve. In addition, the air tube (2) is used to let the air go through, while the bottom of ROSE's skin has a circular-rim shape and is sandwiched by the upper base (4) and the lower base (6) through four bolts (7). This sandwich structure also creates space so that cylinder (5) (fixed to the core of the skin by silicone) can rotate stably at the center of the gripper. The cylinder has a free end that can be assembled with a motor for twist transferring. The inside hole of this cylinder is customized using a square key structure to avoid slipping the skin's silicone axis during rotation. The upper base (6) has three holders to mount additional cameras (3) for vision-based tactile sensing. This design is highly customizable since users may change the actuator according to the working conditions. In this study, we used an external motor and \emph{took advantage of the rotation of the Universal Robot robot arm UR5}\footnote{\url{https://www.universal-robots.com/}} (Teradyne Inc., Denmark) to design different versions of ROSE (see Figure \ref{fabrication}).
\subsection{Fabrication}

In our prototype, the rigid parts were made of polylactide (PLA) plastic while the soft parts were molded by silicone rubber DragonSkin 30 (Smooth-On, Inc. USA). The fabrication process of ROSE gripper is shown in Figure \ref{fabrication}. In the first step, we designed the rigid parts and silicone mold using the 3D CAD software Fusion 360. Next, all rigid parts were fabricated by the 3D printer Sermoon D (Creality 3D, China). Then, silicone rubber was poured into the core molding, flowing over the surface of the core to form the skin shape (with or without markers). This step can be done several times to increase the thickness of the skin. Based on the need of sense, we can fabricate the markers by depositing marker patterns on the core mold. {Here, we manually paint red silicone on the ROSE molds at a series of holes 2\,mm deep and 2\,mm in diameter. After the markers were cured, we cover the mold with white silicone to create the ROSE skin}. Finally, the skin was released from the core mold and assembled with rigid parts and cameras to accomplish a completed ROSE gripper. By applying this process, four versions of ROSE (the non-sensing 4-in, sensing 4-in, non-sensing 2-in, and non-sensing 8-in ones) have been fabricated.

\section{MODELLING}
\label{modelling}
Since ROSE hand enables a large contact area w.r.t. the object, the gripping pressure on the object's surfaces are expected to reduce significantly. At the same time, the strain of the ROSE's thin skin should be clarified w.r.t. the weight of the object. Here, we proposed two simplified models to investigate these factors.
\subsection{Model of Gripping Pressure on Object's Surfaces}
\begin{figure}
    \centering
    \includegraphics[width =0.8\linewidth]{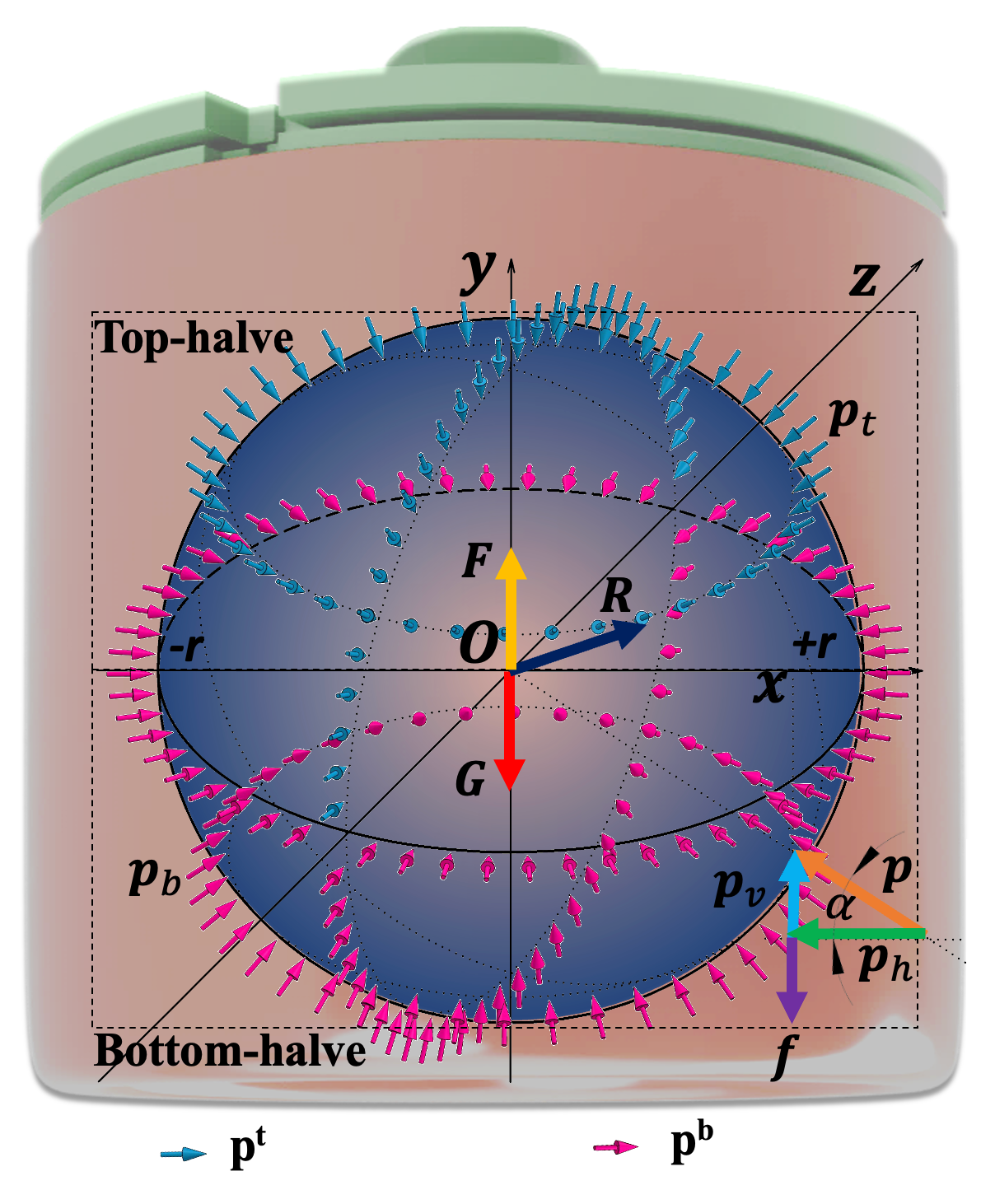}
    \caption{Applied line pressure $p$ on the top-half ($p_t$) and bottom-half ($p_b$) of a spherical object (radius $r$), illustrated by blue and red arrows, respectively. Each pressure consists of the horizontal ($p_h$) and vertical ($p_v$) components at a contact angle $\alpha$.}
    \label{fig: moddeling 1}
\end{figure}
Our model was built based on three assumptions: (1) the gripped object has a spherical shape, (2) the external forces distribute uniformly over each half of a sphere object, {(3) the friction coefficient between the object and ROSE skin $k < 1$, and (4)} the gripped object is covered by at least a bottom half of the ROSE's skin. In this scenario, the external forces applied on the gripped object include the gravity force {\textbf{G}}, friction force $\textbf{F}$, and gripping force $\textbf{R}$ (see Figure \ref{fig: moddeling 1}). Therefore, if we call $\textbf{a}$ and $m$ as the acceleration and the weight of the object, respectively, the dynamic equation of the gripped object can be written as below:
\begin{align}
	m\textbf{a} &= {\textbf{G}} + \textbf{R} + \textbf{F}. 
	\label{eq: 1}
\end{align}
In detail, the gripping and friction forces can be written by defining its distribution at a circular element (diameter at $2x$) of the object as shown in Eqs (\ref{eq 2})-(\ref{eq 3}). Note that while the friction coefficient $\textbf{k}$ (along directions) is considered consistent, the applied pressure on a unit length (\textit{line pressure}) of each half of the object ($\textbf{p}^t$ and $\textbf{p}^b$, respectively) are different. 
\begin{align}
	{\textbf{R}} &= {2}\int_{{0}}^{r} 2 \pi x \textbf{p}^t  \mathrm{d}x + \int_{{0}}^{r} 2 \pi x \textbf{p}^b  \mathrm{d}x \label{eq 2},\\
	{\textbf{F}} &= {2}\int_{0}^{r} 2 \pi x \textbf{p}^t \textbf{.} \textbf{k} \mathrm{d}x + \int_{{0}}^{r} 2 \pi x \textbf{p}^b \textbf{.} \textbf{k} \mathrm{d}x
	\label{eq 3}.
\end{align}
At the equilibrium state ($\textbf{a} = 0$), by projecting all these forces into the vertical direction and using the assumption of a symmetric object, Equation \ref{eq: 1} can be rewritten as
\begin{align}
	m g &= {4}\int_{{0}}^{r} \pi x ({p}^b_v  -  {p}^t_v)  \mathrm{d}x + {4} \int_{{0}}^{r} \pi x ({p}^t_h + {p}^b_h) {k} \mathrm{d}x,\\
	&= 4 \int_{0}^{r} \pi x ({p}^b_v + {p}^b_h k  -  {p}^t_v + {p}^t_h k)  \mathrm{d}x.
	\label{eq 4}
\end{align}
In fact, the line pressure on the top half of an object ($\textbf{p}^t$) is much smaller than that of the bottom half of the object ($\textbf{p}^b$) and $k < 1$ so Equation \ref{eq 4} can be written as
\begin{align}
	m g &= \int_{0}^{r} 4 \pi x ({p}^b_v + {p}^b_h k)  \mathrm{d}x\\
	&= 4 \pi \int_{0}^{r} {p}^b (\sin{\alpha} + k\cos{\alpha} )x \mathrm{d}x.
	\label{eq 5}
\end{align}
Finally, we have
\begin{align}
{p}^b &= \dfrac{m g}{4 \pi \bigints_{\; \;0}^{r} (\dfrac{\sqrt{r^2 - x^2}}{r} + k \dfrac{x}{r}) x \mathrm{d}x},
	\label{eq 6}\\
 {p}^b &= \dfrac{3 m g}{4 \pi (1 + k) r^2}.
	\label{eq 6.1}
\end{align}

With the vertical friction coefficient $k$ defined by practical experiment, the maximum value of applied line pressure $\textbf{p}$ on an object can be estimated using Equation \ref{eq 6.1}. From that, ones can select the materials of ROSE's skin, the size, or the weight of the object to avoid damaging the object.
\subsection{Skin {Vertical} Deformation Calculation}
\label{modelling2}

\begin{figure}
     \centering

        \includegraphics[width = 0.6\linewidth]{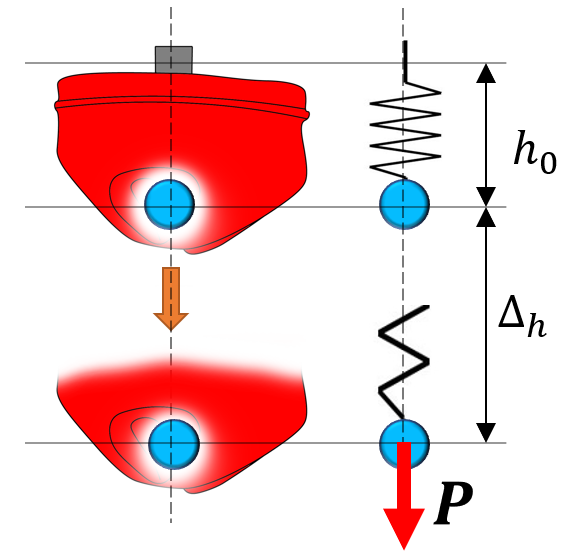}
        \caption{Modelling of ROSE as an equivalent spring. The deformation of the skin is estimated by extension of the spring w.r.t. applied load $P$ of the gripped object.}
        \label{fig: rose_spring}
\end{figure}

\begin{figure*}[ht]
     \centering
     \begin{subfigure}[t]{0.25\linewidth}
        \centering
        \includegraphics[width = \linewidth]{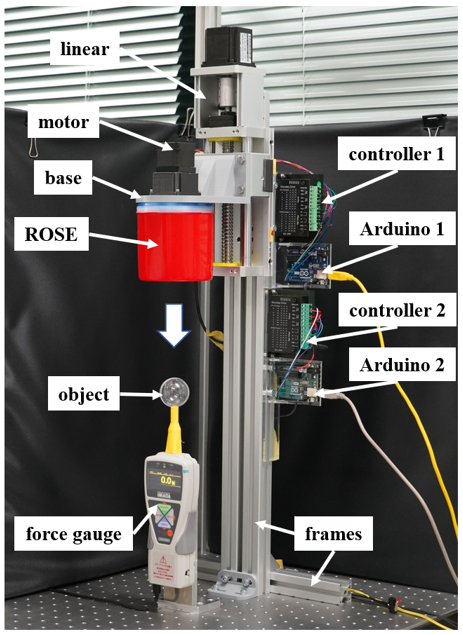}
     \caption{{Experimental setup to measure the payload}}
    \label{fig: rose_payload_setup}
     \end{subfigure}
     \hfill
     \begin{subfigure}[t]{0.73\linewidth}
        \centering
        \includegraphics[width = \linewidth]{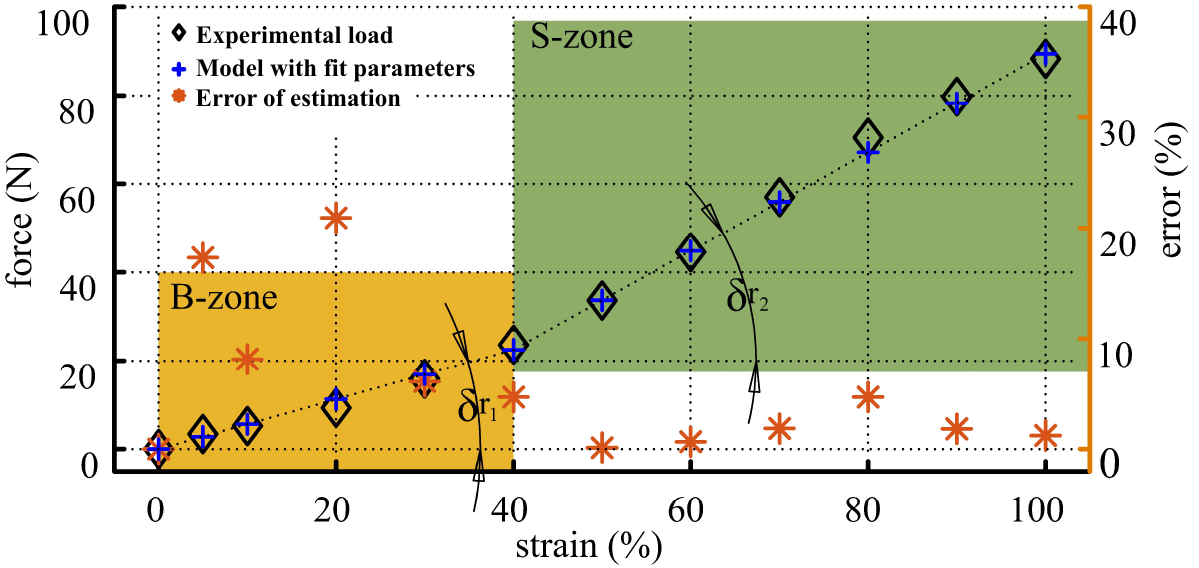}
        \caption{{Comparison of measured gripping force and estimations using the analytical model}}
        \label{fig: rose_payload_result}
     \end{subfigure}
     \hfill
        \caption{{Experiment to evaluate ROSE payload ability. The maximum observed payload was 328.7\,N before the skin is detached from the base (note that the graph is plotted upto 100\,$\%$ strain, thus the maximum value is not recorded here.) }}
        \label{fig: rose_payload_experiment}
\end{figure*}

\par Here, we attempt to estimate the deformation of the {petal} silicone rubber skin while gripping an object, using a simplified analytical approach on a {elastic} spring with a constant cross-section (area $A$). This {assumption} is made to replace the complex morphology of the skin with an uneven distribution of skin thickness in the vertical direction is complex (see Figure \ref{fig: rose_spring}) by a simple elastic spring. Thanks to this assumption, the vertical deformation of ROSE $\Delta_h$ can be defined by the \emph{Hooke}'s law based on the gravity force applied on object $G$ and the stiffness coefficient per area unit of skin $k_0$ as below.
\begin{align}
    \Delta_h = \dfrac{P}{k_0 A},
\end{align}
where $A = V_s/h_0$ with $V_s$ is the volume of ROSE's skin. Then, we obtain:
\begin{align}
    P = \dfrac{k_0 \Delta_h V_s}{h_0}.
    \label{eq:eq11}
\end{align}
Here, we assume to divide the working state of the ROS gripper into two zones: A self-balancing zone (B-zone with equivalent stiffness coefficient of $k_1$) and a stable working zone (S-zone with equivalent stiffness coefficient of $k_2$). Overall, the complex buckling deformation of ROSE's skin leads to uneven strain distribution. In the initial phase (B-zone), the skin starts to twist, resulting in increasing contact area w.r.t object; thus the skin is still soft. When the gripping skin gets maximum, ROSE's skin gets stiffer, implying that the skin's equivalent spring moves to the S-zone with a higher average stiffness coefficient $k_2$. In short, corresponding to each region related to {the mass} of the object, the equivalent spring of the skin behaves differently and is characterized by two coefficients $k_1$ and $k_2$. The Equation \ref{eq:eq11} then can be rewritten as
\begin{align}
    P &= \int_{B-zone} k_1 \dfrac{k_0 V_s}{h_0}\mathrm{d}h + \int_{S-zone} k_2 \dfrac{k_0 V_s}{h_0}\mathrm{d}h.
    \label{eq 12}
\end{align}
\par In detail, $V_s$ can be defined by the geometrical calculation or using the CAD software function regarding the thickness of ROSE's skin; $k_1$ and $k_2$ are practical coefficients and can be obtained emperically. In the same time, from Equation \ref{eq 12}, we can estimate the object's weight by measruing the strain of the ROSE, and vice versa.

\section{Experiment Setup}
\label{experiments}
\subsection{Payload Experiment}

This experiment aims to define the working range of the ROSE, \emph{i.e.} maximum payload. The experiment setup is demonstrated in Figure \ref{fig: rose_payload_setup}. An object was fixed on the top of a vertical ZTA-500N force gauge (IMADA Inc., Japan) through a fixture and connected with this force gauge by an inextensible rope. The ROSE gripper was assembled on a linear state (Suruga Seiki Co., Ltd, Japan) through a 3D-printed base and stayed above and oppositely to the force gauge axis. We used a DC motor 17HS4401S (BIQU Inc., China) as an actuator for rotating the base. The linear stage and the motor were controlled by controller 1 and controller 2, respectively, by two Arduino Rev3 UNO boards (Arudino SRL Inc., America). First, the linear stage moved down to the working position, then ROSE gripped and kept holding this object. Next, the linear stage moved up, resulting in an increase of the vertical force, which resembles increasing the load. {Note that at each strain point data, we stopped the linear stage and rest until the force sensor measuring result is stable for recording. When the strain is over 100\%, we keep moving the linear stage moved up without resting and saved the maximum payload.} The movement of the linear stage and the load were recorded, and processed simultaneously, then shown in Figure \ref{fig: rose_payload_result} and in Table \ref{tab: rose_payload_to_weight_ratio}.
\begin{table}[]
\caption{ROSE's payload-to-weight ratio}
\label{tab: rose_payload_to_weight_ratio}
\resizebox{\linewidth}{!}{%
\begin{tabular}{|l|r|r|r|}
\hline
\textbf{Version} &
  \multicolumn{1}{l|}{\textbf{Weight (kg)}} &
  \multicolumn{1}{l|}{\textbf{\begin{tabular}[c]{@{}l@{}}Maximum\\ payload (kgf)\end{tabular}}} &
  \multicolumn{1}{l|}{\textbf{\begin{tabular}[c]{@{}l@{}}Payload-to-weight\\ ratio\end{tabular}}} \\ \hline
\begin{tabular}[c]{@{}l@{}}Payload test\\ (ROSE + rotor)\end{tabular} &
  0.491 &
  33.51 &
  6812\% \\ \hline
\end{tabular}%
}
\end{table}
\subsection{Sensing Experiment}
\begin{figure}[ht]
     \centering
     \begin{subfigure}[t]{0.49\linewidth}
        \centering
        \includegraphics[width = \linewidth]{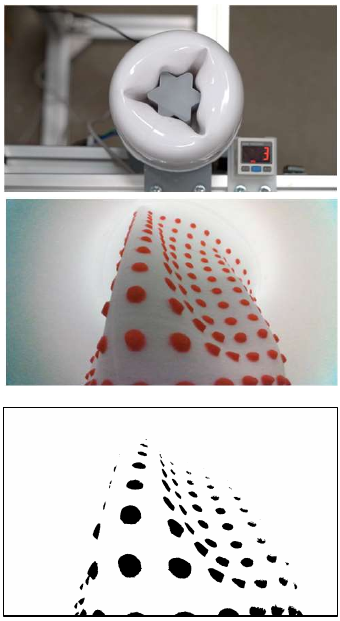}
        \caption{ROSE with air support at the start frame}
        \label{fig: rose_with_air_start_frame}
     \end{subfigure}
     \hfill
     \begin{subfigure}[t]{0.49\linewidth}
        \centering
        \includegraphics[width = \linewidth]{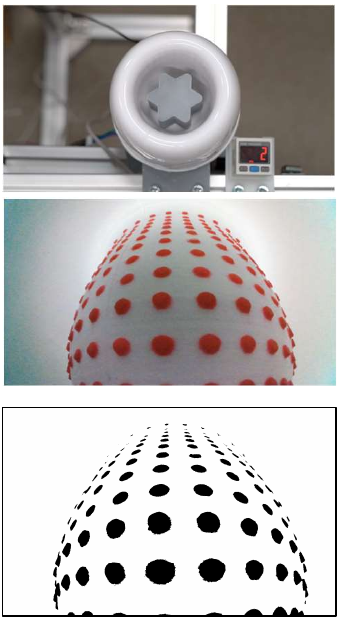}
        \caption{ROSE without air support at the start frame}
        \label{fig: rose_without_air_start_frame}
     \end{subfigure}
     \hfill
     \hfill
        \caption{ROSE's skin with inner markers and related tactile sensing experiment setup (with/without air support). From top row to bottom row: top view, inside camera view, and binary frame processed by OpenCV.}
        \label{fig: rose_sensing_experiment}
\end{figure}
In this experiment, we used a white skin for the ROSE gripper version, with red markers at the inner wall. We integrated a color cameras (MISUMI Electronics Corp., Taiwan) for monitoring the movements of markers during the operation of the ROSE. Objects were put into the gripper and then gripped with and without air support at 3\,kPa. Here, the air support was inputted inside the ROSE's skin to vary the stiffness of the inflated skin, also helping better embracement of the skin to the gripped object. The air was controlled by a solenoid valve VQD 1121-5L (SMC Corp., Japan) and a pressure sensor ISE30A-C6H-N-M (SMC Corp., Japan). A camera DSC-RX10M4 (Sony, Japan) was set outside and above the ROSE to record the gripping behavior while the integrated cameras capture the movement of the skin and markers. Finally, we processed all data from the integrated cameras by OpenCV in Python to evaluate the sensing potential of ROSE (see Figure \ref{fig: rose_sensing_experiment}) and the video\footnote{\url{https://youtu.be/E1wAI09LaoY}}.
\subsection{Durability Experiment}
\label{sec: durability_experiment}
\begin{figure}
     \centering
     \begin{subfigure}[t]{0.40\linewidth}
        \centering
      \includegraphics[width=\textwidth]{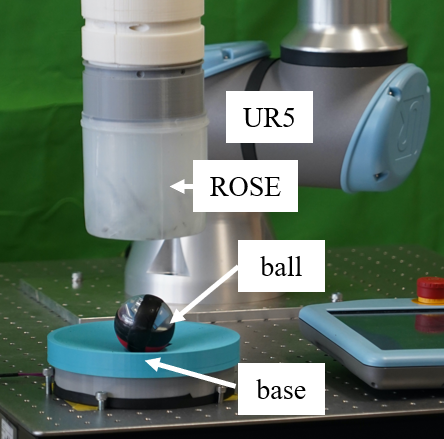}
       \caption{Durability test setup}
      \label{fig: durability_test}
     \end{subfigure}
     \hfill
      \begin{subfigure}[t]{0.58\linewidth}
    \centering
      \includegraphics[width=\textwidth]{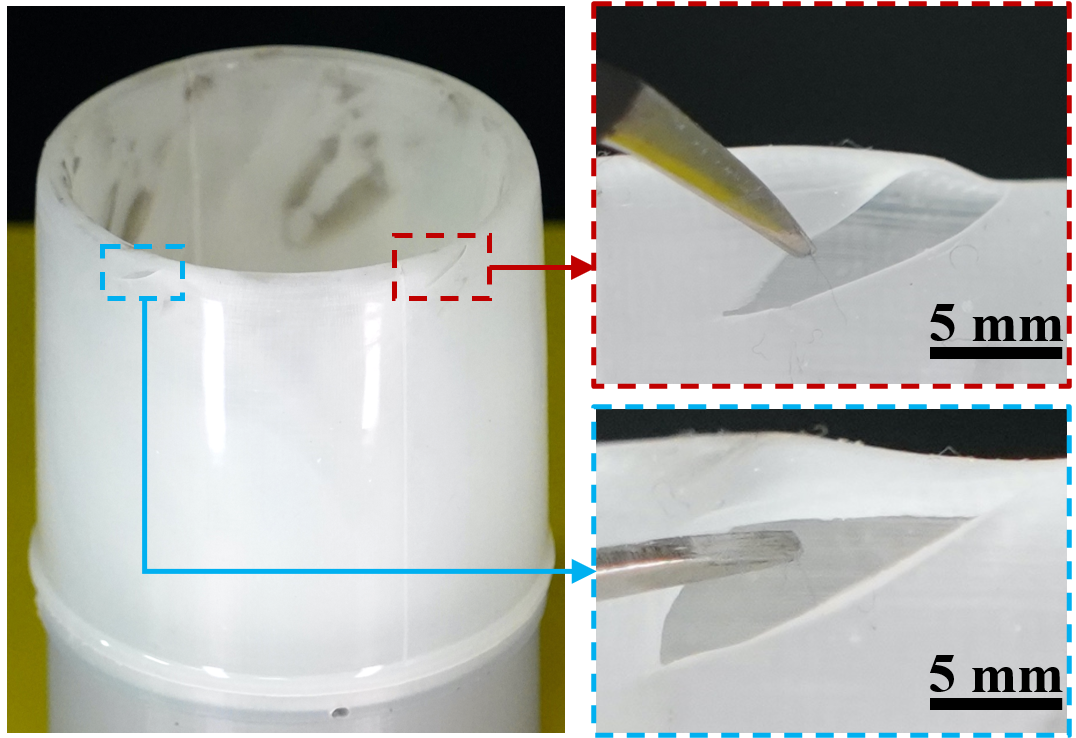}
       \caption{Durability test result after $400,000$ times of open-close trials}
      \label{fig: damage_test_1}
     \end{subfigure}
     \begin{subfigure}[t]{\linewidth}
        \centering
      \includegraphics[width=\textwidth]{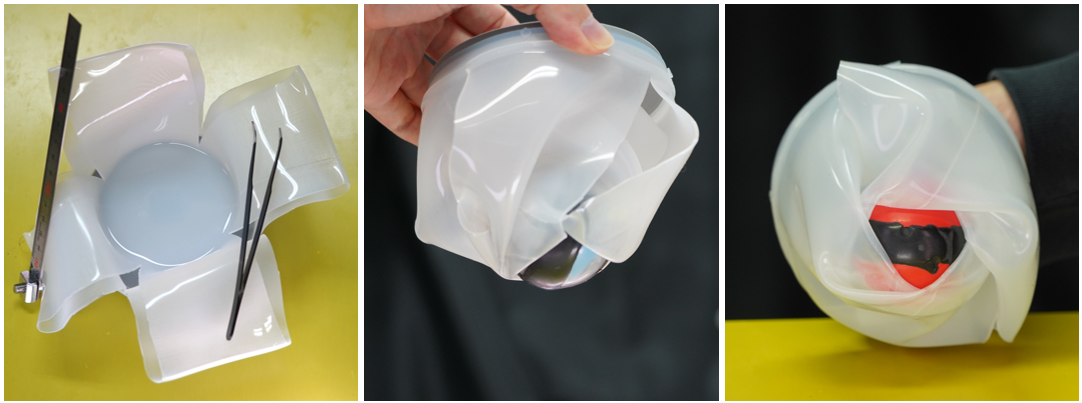}
       \caption{Durability test result in case of critical damage ({4 cut lines at 100\% of skin height}), from left to right: broken ROSE, front view, and bottom view of a broken ROSE successful grasping}
      \label{fig: damage_test_2}
     \end{subfigure}
     \hfill
        \caption{Durability test: setup and result}
        \label{fig: durability_experiment}
\end{figure}

Durability problem is always raised for soft robotic hands. Here, we attempt to demonstrate the ROSE's durability through two cases: (1) repetitive grasping task, and (2) critical damage. In this experiment, we used the 4-inch ROSE to grasp a $210$\,g, $50$\,mm-diameter ball (similar to a typical apple), made by gluing two hemispheres by a black tape (see Figure \ref{fig: durability_test}). 
\par In the first test, UR5 was used to repeat the ROSE's grasp, hold, and release the object (see Figure \ref{fig: damage_test_1}). After every 24 hours, we stopped the UR5 to check ROSE and saved the data to a computer.
\par For the second test, we cut the inner and the outer skins of the ROSE by $12.5, 25, 37.5, 50, 62.5, 75, 87.5, 100 \%$ of their height and check when it would fail to grasp objects (see Figure \ref{fig: damage_test_2}).
\subsection{Demonstrations with UR5}
\begin{figure*}[ht]
      \centering
      \includegraphics[width=\textwidth]{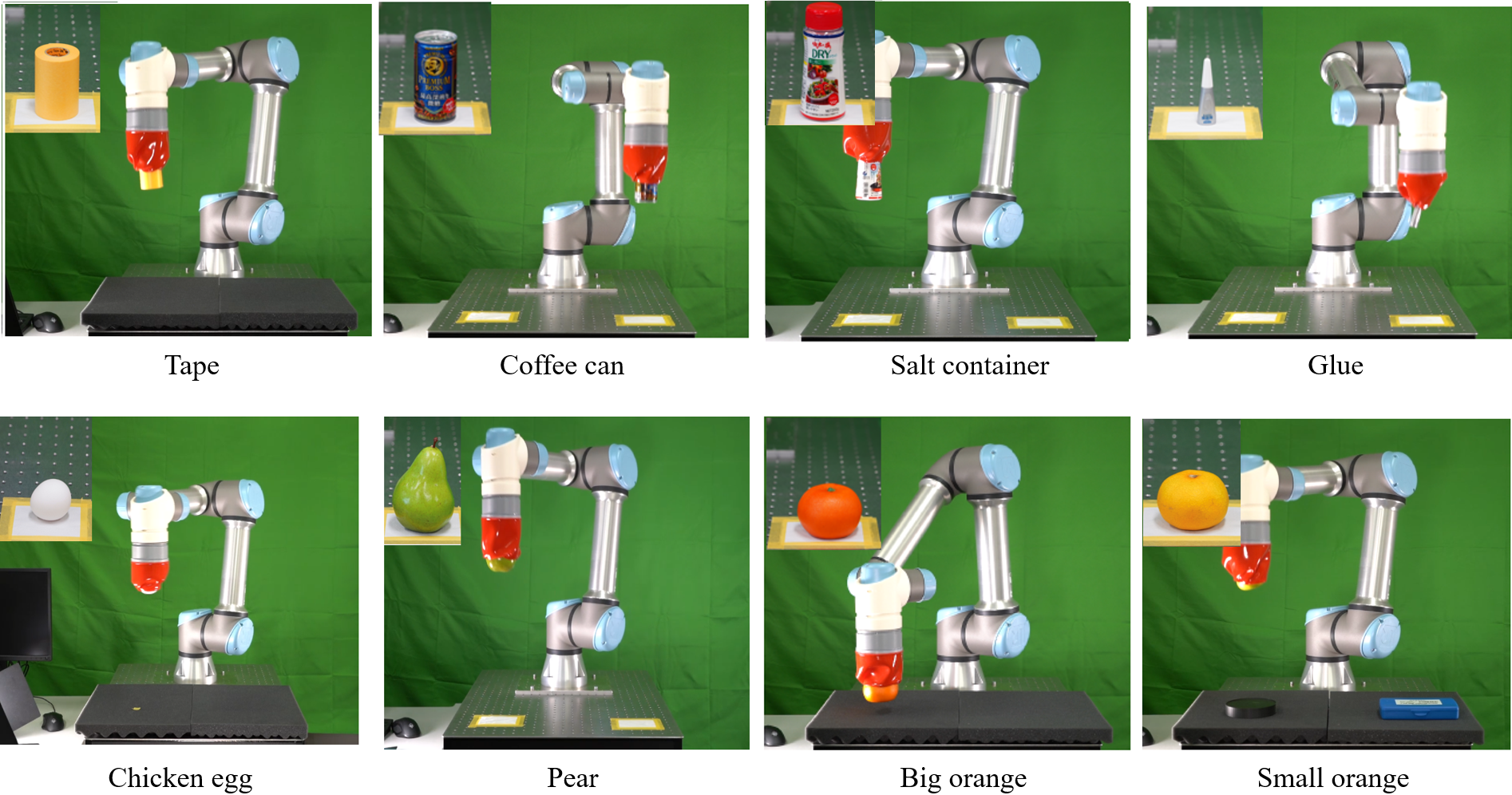}
       \caption{Demonstration of ROSE on UR5 grasping objects with various sizes, shapes, and materials}
      \label{fig: demonstration}
  \end{figure*}
\begin{table*}[ht]
\caption{Demonstration of ROSE on UR5 with various size, shape and materials objects {after 30 times trials}}
\label{tab: demonstration}
\centering
\resizebox{\linewidth}{!}{%
\begin{tabular}{lllllllll} 
\hline
Object & Tape & Coffee can & Salt container & Glue & Chicken egg & Pear & Big orange & Small orange \\ 
\hline
Weight [g] & 93 & 216 & 235 & 12 & 57 & 245 & 120 & 96 \\
Size (H x D) [mm] & 60 x 52 & 105 x 53 & 238 x 56 & 85 x 23 & 42 x 60 & 100 x 79 & 60 x 79 & 50 x 62 \\
Materials & Fabrics & Steel & Plastics & Plastics & Raw egg & Organic & Organic & Organic \\
Success rate & \textbf{100\%} & \textbf{100\%} & \textbf{100\%} & \textbf{100\%} & \textbf{100\%} & \textbf{100\%} & \textbf{100\%} & \textbf{100\%} \\
\hline
\end{tabular}
}
\end{table*}
  Several common objects such as fruits, a coffee can, and a chicken egg were used to demonstrate the gripping ability of ROSE by using a UR5 robot arm. The ROSE gripper was assembled directly to the final joint of this robot arm, and the trajectories of this arm were programmed manually from the control panel. The result has summarised in Table \ref{tab: demonstration} and illustrated in Figure \ref{fig: demonstration}. The performance of ROSE was recorded, and detail of this experiment can be found in our video. In order to see the performance of ROSE with wet and deformable objects, a challenging mission for robotic grasping and manipulation, we designed a critical experiment with a peeled boiled egg (45\,mm in diameter and 52\,mm in height). Here, a UR5 robot arm equipped with ROSE was used to grasp this egg in an olive oil container with different percentages of submerging (0, 30, 60, and 90\%). Finally, the results of these grasping demonstrations are summarized in Table \ref{tab: rose_olive_egg}, Figure \ref{fig: rose_introduction}, and our video$^3$.
  
\begin{table}[]
\caption{Result of grasping peeled boiled chicken egg submerged inside an olive oil's container}
\label{tab: rose_olive_egg}
\resizebox{\linewidth}{!}{%
\begin{tabular}{|l|l|l|l|l|}
\hline
Percentage of submerge & 0\%       & 30\%      & 60\%      & 90\%                             \\ \hline
Success rate          & \textbf{100\%} & \textbf{100\%} & 8\% & {0\%} \\ \hline
\end{tabular}%
}
\end{table}
\section{Results}
\label{results}
\subsection{Payload}
The payload experiment indicates that ROSE could endure the maximum payload at 328.7\,N generated by 49\,g silicone skin, thanks to the morphological change of the skin, indicated by the contact angle $\alpha$ in Equation \ref{eq 5}, which also describes that the larger $\alpha$ that ROSE has, the lower the grasping pressure is. Therefore, with the close structure, ROSE offers a high $\alpha$ (90 degrees at the locking base). This locking property explains why ROSE can reach a high lift force. As a result, ROSE gripper has a lofty payload-to-weight ratio of about 6812\% for the gripper alone (without the twisting motor).
\par Figure \ref{fig: rose_payload_experiment} show the generated force of ROSE's skin w.r.t the various payload. We also attempted to compare the estimation using Eq. \ref{eq 12}. The emperical stiff coefficients $k_1$ and $k_2$ of the equivalent spring were interpolated from the payload experiment result. Here, the model showed good strain prediction when the errors of simulation mostly are under 10\,\%. Secondly, the transition between the B-zone and S-zone as mentioned in section \ref{modelling2} is clearly seen at the payload around 40\,N. It also implies that the more applied load, the harder the skin performs. This resembles to the assumption of the skin cover increasing the grasping process in section \ref{modelling2}.

 \subsection{Gripping Ability}
Our demonstration shows that ROSE can gently manipulate various sizes, shapes, and materials objects without any damage or crash, including food products (see Fig. \ref{fig: demonstration}) with 100\,\% success. During this experiment, ROSE showed good stability while gripping, transporting, and placing these objects. Additionally, ROSE can pick up other small objects such as food (peanuts, rice, or candy), mechanical items (bolts, screws, nuts), and granular media such as gravel.
\par Especially, the success of grasping peeled egg partly sub-submerged inside an oil container (see Table \ref{tab: rose_olive_egg}) clarifies that ROSE can also work in a critical wet, slippery, and deformable object that other conventional soft robotics grippers could not accomplish (to best of our knowledge). The failure to grasp at $42\,mm$ height of olive oil (fully sub-submerged underneath the oil surface) is due to the remaining air in ROSE's hand during approaching the egg inside the oil container, preventing the egg from sucking into the pedal skin of ROSE. In fact, we have changed the gripping strategy by combining the approaching movement and rotation of ROSE's pedal skin to squeeze out the air. As a result, the strategy for efficiently gripping objects needs to be elaborated in depth in the future.

\subsection{Durability}
\par After 400,000 times of grasping, ROSE's skin has some small tears as observed in Figure \ref{fig: damage_test_1}. This result indicates that ROSE gripper can work durably thank to its simple but efficient structure. Especially, such small tears on the ROSE gripper's skin do not affect the working ability of ROSE in our experiment. Therefore, in order to verify the limitation of ROSE, we cut 100\,\% of its length at four symmetric directions (Figure \ref{fig: damage_test_1}), and found that ROSE still could keep grasping successfully. This is considered to be significant since most of the soft robotic hands (without self-healing) could not perform the task upon being partly damaged. 
\subsection{Sensing Ability}
In this paper, a design of vision-based tactile sensing using markers was proposed. Based on the preliminary result obtained from the cameras, shows potential in estimating the size and shape, as well as states of gripped objects. Figures \ref{fig: rose_with_air_start_frame}-\ref{fig: rose_without_air_start_frame} indicate clearly the difference in terms of markers' distribution, thus the skin's condition, upon contact with the object. Even though perception and feedback control using tactile images are not the focus of this paper, rich information from markers' movement and distribution shown in this experiment promises the necessary database for learning states of contact. It is also learned from the preliminary results that buckling of the skin upon contact with an object causes occlusions of markers, which may be overcome by a combination of images from three cameras distributed equally around the base and learning algorithm. More elaborations on the tactile perception of the ROSE gripper will be reported in the future.

\section{Discussion and Conclusion}
\label{discussion_and_conlcusion}
\subsection{Design and Fabrication}
The design of ROSE is novel but simple, for example, the non-sensing 4-in version for the UR5 robot arm requires only five components (3D-printed parts and a skin) and four pairs of bolts and nuts, which has a total weight under $200\,g$ and supports plug and play ability (see the video). This results in ease of fabrication. While rigid parts can be 3D-printed and ROSE's skin can fast fabricated with a simple one-core mold; this fabrication method challenges the control of ROSE's skin thickness. A solution could be multi-cavity injection molds, which are expensive but promising for mass production with consistent thickness. 

\par The size, shape, or sensing ability of ROSE can be highly customized thanks to this simple design. Since robot arms today often have a rotational DoF at the position of the end effector, ROSE gripper can be easily customized to connect with an industrial arm. As a result, none of the extra rotors is needed to activate ROSE gripper. Additionally, assembled by 3D-printed PLA parts and silicones, a customized ROSE version for commercial robot arm UR5 reduces significantly the cost of the gripper.

\subsection{Gripping Ability}
Our experiment results reveal that objects should fall within the space bounded by the {petal} skin. Thanks to the symmetric design and squeezing mechanism, ROSE gripper can embrace objects without precise alignment w.r.t. the object. This ability suggests that ROSE can handle objects with less burden in computation (for controlling the gripper's position and orientation). Also, there is no problem when the gripper hits the ground (table) while approaching the object, thanks to the {softness} of the pental skin. Regardless this, ROSE gripper meets difficulties in handling oversized objects (in comparison with ROSE's size). ROSE fails to grasp flat objects such as CDs, or elongated objects such as poles. However, it can grip a piece of fabric at specific postures since the twist action can suck part of the fabric into a squeeze, thus successfully lifting it. The ROSE gripper, thus, may need several renovations for handling such objects as future work. 

For grasping objects inside a liquid container, the structure of the skin needs to be changed by integrating the air leaked mechanism in combination with a control strategy for gripping successfully in this scenario. Moreover, the air support for sensing opens up the idea of taking advantage of a pneumatic system to support the holding step and quick releasing. Such work is also left as future elaboration.

\subsection{Modelling}

The large contact area and buckling structure help ROSE apply low pressure on the grasped object surface but still endure high payload. The former advantage can be seen in Equation \ref{eq 6.1} of our first model which indicates that the applied pressure of ROSE over object weight is approximately linear. Regardless, such a simplified model only helps with the preliminary evaluation of the grippers' design and operation. In fact, the buckling deformation of the ROSE skin during the grasping process is nonlinear and complex. While our simplified models can explain the behavior and trend of the payload, lacking gripping pressure model evaluation is a shortage, even though it is not the main focus of this paper. In the future, we seek for total solution in detailed modeling of the mechanical behavior of the ROSE's skin since it is also helpful for the creation of the ROSE's digital twin for embedding in robotics simulation tools. A digital twin is also applicable for data acquisition of tactile images, which then leads to the creation of tactile perception for ROSE through learning methods.

\subsection{Applications}
These novel gripper ROSEs are anticipated to be used on robot arms for manipulating tasks. With high durability, ROSE ripper can be applied to tasks that require continuity for a long time such as classification at factories and agricultural harvesting on farm (where the postures of fruits are diverse). It also brings a solution for the seafood industry where modified ROSE can grasp underwater products. The close structure of ROSE suggests an application for other fields such as packaging where the skin can cover items and remove this combo from ROSE's base as a package. In nearly future, we believe that the pictures of ROSE in restaurants or factories would be feasible. Moreover, we would also like to use bio-degradable materials to fabricate ROSE's skin toward a sustainable solution for nature.
\section{Conclusion}
\label{conclusion}
In this article, we proposed the design and fabrication of a novel bio-inspired gripper named ROSE. An examination of ROSE's characteristics clarified its features including the maximum payload, payload-to-weight ratio, durability, skin deformation, and contact perception. In addition, a demonstration with multiple objects, including an shelled chicken egg dumped in olive oil, clarifies the universal grasping ability of ROSE. In the future, we plan to simulate and evaluate the behavior of ROSE's skin with specialized software and practical experiments, respectively. Furthermore, we intend to develop the design of ROSE and explore its applications in other scenarios such as underwater grasping. 
\bibliographystyle{IEEEtran}
\bibliography{main}

\begin{thebibliography}{10}
\providecommand{\url}[1]{#1}
\csname url@samestyle\endcsname
\providecommand{\newblock}{\relax}
\providecommand{\bibinfo}[2]{#2}
\providecommand{\BIBentrySTDinterwordspacing}{\spaceskip=0pt\relax}
\providecommand{\BIBentryALTinterwordstretchfactor}{4}
\providecommand{\BIBentryALTinterwordspacing}{\spaceskip=\fontdimen2\font plus
\BIBentryALTinterwordstretchfactor\fontdimen3\font minus
  \fontdimen4\font\relax}
\providecommand{\BIBforeignlanguage}[2]{{%
\expandafter\ifx\csname l@#1\endcsname\relax
\typeout{** WARNING: IEEEtran.bst: No hyphenation pattern has been}%
\typeout{** loaded for the language `#1'. Using the pattern for}%
\typeout{** the default language instead.}%
\else
\language=\csname l@#1\endcsname
\fi
#2}}
\providecommand{\BIBdecl}{\relax}
\BIBdecl

\bibitem{finger_1}
E.~Turco, V.~Bo, M.~Pozzi, A.~Rizzo, and D.~Prattichizzo, ``Grasp planning with
  a soft reconfigurable gripper exploiting embedded and environmental
  constraints,'' \emph{IEEE Robotics and Automation Letters}, vol.~6, no.~3,
  pp. 5215--5222, 2021.

\bibitem{octopus_suction_1}
T.~Takahashi, M.~Suzuki, and S.~Aoyagi, ``Octopus bioinspired vacuum gripper
  with micro bumps,'' in \emph{2016 IEEE 11th Annual International Conference
  on Nano/Micro Engineered and Molecular Systems (NEMS)}, 2016, pp. 508--511.

\bibitem{flytrap}
L.~Xu and G.~Gu, ``Bioinspired venus flytrap : A dielectric elastomer actuated
  soft gripper,'' in \emph{2017 24th International Conference on Mechatronics
  and Machine Vision in Practice (M2VIP)}, 2017, pp. 1--3.

\bibitem{octopus_suction_2}
\BIBentryALTinterwordspacing
M.~Wu, X.~Zheng, R.~Liu, N.~Hou, W.~H. Afridi, R.~H. Afridi, X.~Guo, J.~Wu,
  C.~Wang, and G.~Xie, ``Glowing sucker octopus (stauroteuthis
  syrtensis)-inspired soft robotic gripper for underwater self-adaptive
  grasping and sensing,'' \emph{Advanced Science}, vol.~9, no.~17, p. 2104382,
  2021. [Online]. Available:
  \url{https://onlinelibrary.wiley.com/doi/abs/10.1002/advs.202104382}
\BIBentrySTDinterwordspacing

\bibitem{elephenttrunk1}
\BIBentryALTinterwordspacing
Y.~Li, Y.~Chen, T.~Ren, Y.~Li, and S.~h. Choi, ``Precharged pneumatic soft
  actuators and their applications to untethered soft robots,'' \emph{Soft
  Robotics}, vol.~5, no.~5, pp. 567--575, 2018, pMID: 29924683. [Online].
  Available: \url{https://doi.org/10.1089/soro.2017.0090}
\BIBentrySTDinterwordspacing

\bibitem{elephenttrunk2}
M.~E. Giannaccini, I.~Georgilas, I.~Horsfield, B.~H. Peiris, A.~Lenz, A.~G.
  Pipe, and S.~Dogramadzi, ``A variable compliance, soft gripper,''
  \emph{Autonomous Robots}, vol.~36, pp. 93--107, 2014.

\bibitem{perching}
\BIBentryALTinterwordspacing
W.~R.~T. Roderick, M.~R. Cutkosky, and D.~Lentink, ``Bird-inspired dynamic
  grasping and perching in arboreal environments,'' \emph{Science Robotics},
  vol.~6, no.~61, p. eabj7562, 2021. [Online]. Available:
  \url{https://www.science.org/doi/abs/10.1126/scirobotics.abj7562}
\BIBentrySTDinterwordspacing

\bibitem{flowers1}
\BIBentryALTinterwordspacing
Z.~Zhang, X.~Ni, W.~Gao, H.~Shen, M.~Sun, G.~Guo, H.~Wu, and S.~Jiang,
  ``Pneumatically controlled reconfigurable bistable bionic flower for robotic
  gripper,'' \emph{Soft Robotics}, vol.~9, no.~4, pp. 657--668, 2022, pMID:
  34287072. [Online]. Available: \url{https://doi.org/10.1089/soro.2020.0200}
\BIBentrySTDinterwordspacing

\bibitem{flowers2}
\BIBentryALTinterwordspacing
F.~Hu, L.~Lyu, and Y.~He, ``A 3d printed paper-based thermally driven soft
  robotic gripper inspired by cabbage,'' \emph{International Journal of
  Precision Engineering and Manufacturing}, vol.~20, pp. 1915--1928, 2019.
  [Online]. Available: \url{https://doi.org/10.1007/s12541-019-00199-6}
\BIBentrySTDinterwordspacing

\bibitem{CAGE}
\BIBentryALTinterwordspacing
S.~Makita and W.~Wan, ``A survey of robotic caging and its applications,''
  \emph{Advanced Robotics}, vol.~31, no. 19-20, pp. 1071--1085, 2017. [Online].
  Available: \url{https://doi.org/10.1080/01691864.2017.1371075}
\BIBentrySTDinterwordspacing

\bibitem{WRAP}
V.~A. Ho, ``Grasping by wrapping: Mechanical design and evaluation,'' in
  \emph{2017 IEEE/RSJ International Conference on Intelligent Robots and
  Systems (IROS)}, 2017, pp. 6013--6019.

\bibitem{roadmap}
\BIBentryALTinterwordspacing
B.~Mazzolai \emph{et~al.}, ``Roadmap on soft robotics: multifunctionality,
  adaptability and growth without borders,'' \emph{Multifunctional Materials},
  vol.~5, no.~3, p. 032001, aug 2022. [Online]. Available:
  \url{https://dx.doi.org/10.1088/2399-7532/ac4c95}
\BIBentrySTDinterwordspacing

\bibitem{2fingers}
S.~Liu, F.~Wang, Z.~Liu, W.~Zhang, Y.~Tian, and D.~Zhang, ``A two-finger
  soft-robotic gripper with enveloping and pinching grasping modes,''
  \emph{IEEE/ASME Transactions on Mechatronics}, vol.~26, no.~1, pp. 146--155,
  2021.

\bibitem{finger_palm}
A.~Pagoli, F.~Chapelle, J.~A. Corrales, Y.~Mezouar, and Y.~Lapusta, ``A soft
  robotic gripper with an active palm and reconfigurable fingers for fully
  dexterous in-hand manipulation,'' \emph{IEEE Robotics and Automation
  Letters}, vol.~6, no.~4, pp. 7706--7713, 2021.

\bibitem{4fingers}
W.~Hu and G.~Alici, ``Bioinspired three-dimensional-printed helical soft
  pneumatic actuators and their characterization,'' \emph{Soft Robotics},
  vol.~7, pp. 267--282, 2020.

\bibitem{tofugripper}
P.~Van~Nguyen, Q.~K. Luu, Y.~Takamura, and V.~A. Ho, ``Wet adhesion of
  micro-patterned interfaces for stable grasping of deformable objects,'' in
  \emph{2020 IEEE/RSJ International Conference on Intelligent Robots and
  Systems (IROS)}, 2020, pp. 9213--9219.

\bibitem{magicball}
S.~Li, J.~J. Stampfli, H.~J. Xu, E.~Malkin, E.~V. Diaz, D.~Rus, and R.~J. Wood,
  ``A vacuum-driven origami “magic-ball” soft gripper,'' in \emph{2019
  International Conference on Robotics and Automation (ICRA)}, 2019, pp.
  7401--7408.

\bibitem{Haili2023}
L.~I. Haili, Z.~Shuai, Z.~Xuanhao, Z.~Wumian, and Y.~A.~O. Jiantao, ``A 0 .
  5-meter-scale, high-load, soft-enclosed gripper capable of grasping the human
  body,'' \emph{Sci. China Technol. Sci.}, 2023.

\bibitem{Li2021b}
\BIBentryALTinterwordspacing
H.~Li, J.~Yao, C.~Wei, P.~Zhou, Y.~Xu, and Y.~Zhao, ``An untethered soft
  robotic gripper with high payload-to-weight ratio,'' \emph{Mechanism and
  Machine Theory}, vol. 158, p. 104226, 2021. [Online]. Available:
  \url{https://doi.org/10.1016/j.mechmachtheory.2020.104226}
\BIBentrySTDinterwordspacing

\bibitem{Pedro2018}
P.~Pedro, C.~Ananda, P.~B. Rafael, A.~R. Carlos, and B.~C. Alexandre, ``Closed
  structure soft robotic gripper,'' \emph{2018 IEEE International Conference on
  Soft Robotics, RoboSoft 2018}, vol.~2, pp. 66--70, 2018.

\bibitem{9200339}
Y.~Hao, S.~Biswas, E.~W. Hawkes, T.~Wang, M.~Zhu, L.~Wen, and Y.~Visell, ``A
  multimodal, enveloping soft gripper: Shape conformation, bioinspired
  adhesion, and expansion-driven suction,'' \emph{IEEE Transactions on
  Robotics}, vol.~37, no.~2, pp. 350--362, 2021.

\bibitem{jamminggripper1}
J.~R. Amend, E.~Brown, N.~Rodenberg, H.~M. Jaeger, and H.~Lipson, ``A positive
  pressure universal gripper based on the jamming of granular material,''
  \emph{IEEE Transactions on Robotics}, vol.~28, no.~2, pp. 341--350, 2012.

\bibitem{jamming2}
\BIBentryALTinterwordspacing
S.~D’Avella, P.~Tripicchio, and C.~A. Avizzano, ``A study on picking objects
  in cluttered environments: Exploiting depth features for a custom low-cost
  universal jamming gripper,'' \emph{Robotics and Computer-Integrated
  Manufacturing}, vol.~63, p. 101888, 2020. [Online]. Available:
  \url{https://www.sciencedirect.com/science/article/pii/S0736584519307276}
\BIBentrySTDinterwordspacing

\bibitem{seasampling}
\BIBentryALTinterwordspacing
S.~Licht, E.~Collins, M.~L. Mendes, and C.~Baxter, ``Stronger at depth: Jamming
  grippers as deep sea sampling tools,'' \emph{Soft Robotics}, vol.~4, no.~4,
  pp. 305--316, 2017, pMID: 29251570. [Online]. Available:
  \url{https://doi.org/10.1089/soro.2017.0028}
\BIBentrySTDinterwordspacing

\bibitem{softbuble}
N.~Kuppuswamy, A.~Alspach, A.~Uttamchandani, S.~Creasey, T.~Ikeda, and
  R.~Tedrake, ``Soft-bubble grippers for robust and perceptive manipulation,''
  in \emph{2020 IEEE/RSJ International Conference on Intelligent Robots and
  Systems (IROS)}, 2020, pp. 9917--9924.

\bibitem{scooplinggripper}
\BIBentryALTinterwordspacing
Z.~Wang, H.~Furuta, S.~Hirai, and S.~Kawamura, ``A scooping-binding robotic
  gripper for handling various food products,'' \emph{Frontiers in Robotics and
  AI}, vol.~8, 2021. [Online]. Available:
  \url{https://www.frontiersin.org/articles/10.3389/frobt.2021.640805}
\BIBentrySTDinterwordspacing

\end{thebibliography}
\end{document}